\documentclass{article}

\usepackage{arxiv}

\usepackage{cite}
\usepackage{amsmath,amssymb,amsfonts}
\usepackage{algorithmic}
\usepackage{graphicx}
\usepackage{textcomp}
\usepackage{xcolor}
\usepackage{booktabs}
\usepackage{colortbl}
\usepackage{bm}
\usepackage{subcaption}
\usepackage{multirow}
\usepackage{multicol}

\usepackage[ruled,linesnumbered]{algorithm2e}
\DeclareMathOperator*{\argmax}{argmax}
\DeclareMathOperator*{\argmin}{argmin}

\def\opA{\mathfrak{A}}
\def\art{\mathfrak{a}}

\def\caLt{\mathcal{L}_\text{target}}

\def\naDt{\nabla_\text{target}}

\def\mask{\mathbf{m}}

\def\BibTeX{{\rm B\kern-.05em{\sc i\kern-.025em b}\kern-.08em
    T\kern-.1667em\lower.7ex\hbox{E}\kern-.125emX}}

\renewcommand{\shorttitle}{NSA: Naturalistic Support Artifact to Boost Network Confidence}

\title{NSA: Naturalistic Support Artifact to Boost Network Confidence
\thanks{\textit{Accepted at IEEE International Joint Conference on Neural Networks 2023} } 
}

\author{Abhijith Sharma \\
Department of Computer Science \\
University of British Columbia\\
BC, Canada \\
\texttt{sharma86@mail.ubc.ca}
\And
Phil Munz \\
TrojAI Inc. \\
NB, Canada \\
\texttt{phil.munz@troj.ai}
\And
Apurva Narayan\\
Department of Computer Science \\
Western University\\
ON, Canada \\
\texttt{apurva.narayan@uwo.ca}
}

\begin{document}
\maketitle

\begin{abstract}
Visual AI systems are vulnerable to natural and synthetic physical corruption in the real-world. Such corruption often arises unexpectedly and alters the model's performance. In recent years, the primary focus has been on adversarial attacks. However, natural corruptions (e.g., snow, fog, dust) are an omnipresent threat to visual AI systems and should be considered equally important. Many existing works propose interesting solutions to train robust models against natural corruption. These works either leverage image augmentations,  which come with the additional cost of model training, or place suspicious patches in the scene to design unadversarial examples. In this work, we propose the idea of naturalistic support artifacts (NSA) for robust prediction. The NSAs are shown to be beneficial in scenarios where model parameters are inaccessible and adding artifacts in the scene is feasible. The NSAs are natural looking objects generated through artifact training using DC-GAN to have high visual fidelity in the scene. We test against natural corruptions on the Imagenette dataset and observe the improvement in prediction confidence score by four times. We also demonstrate NSA's capability to increase adversarial accuracy by 8\% on average. Lastly, we qualitatively analyze NSAs using saliency maps to understand how they help improve prediction confidence. 
\end{abstract}


\section{Introduction}
Image processing has become indispensable to most vision-based applications in recent years. Simultaneously, convolutional neural networks (CNNs) have gained traction due to their ability to handle visual inputs and achieve human-level performance for specific tasks \cite{taigman2014deepface}, \cite{mnih2015human}.
Eventually, CNNs found their way into numerous applications for scene understanding, and automated detection of objects \cite{gu2019survey}, \cite{guo2021survey}. The emergence of CNNs has been remarkable, but their performance is highly conditioned on their prior training distribution \cite{fang2020rethinking}. The samples from out-of-distribution have led to erroneous predictions \cite{hendrycks2021many}, \cite{hendrycks2019benchmarking}, even failing miserably in some scenarios \cite{kopestinsky25astonishing}. This led to a series of works that have concentrated on inspecting the fragile behavior of neural networks \cite{madry2017towards}, \cite{ilyas2019adversarial}, \cite{szegedy2013intriguing} and the natural trade-off between accuracy and robustness \cite{tsipras2018robustness}. The robustness aspects of an AI model have become just as necessary as traditional accuracy metrics, especially for safety-critical systems. 

 \begin{figure*}[t]
     \centering 
     \includegraphics[scale=0.56, angle=270]{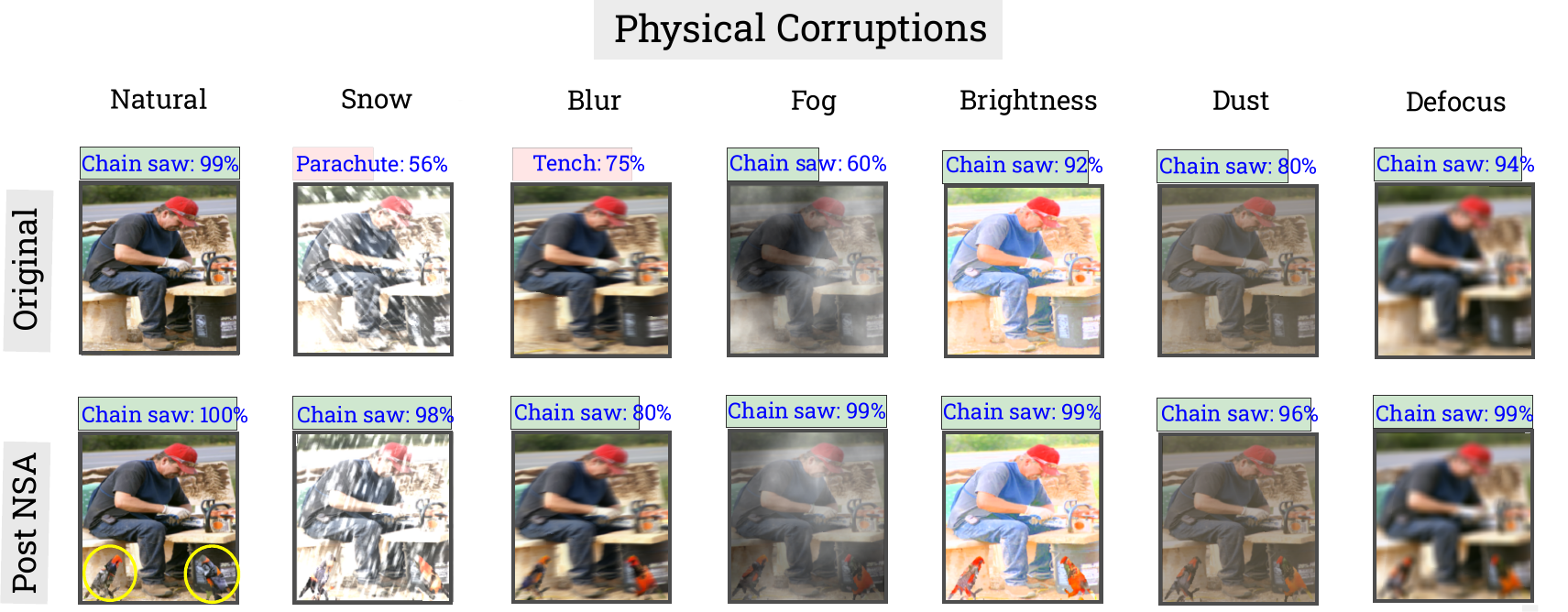}
     \caption{
     The figure shows the increase in the prediction score after placing NSA for various natural corruptions. The original prediction is Chain Saw. The top row is without NSA (original image), and the bottom row is with NSA. The yellow circle highlights the presence of artifacts (NSA). We use two bird artifacts as they have high visual fidelity in the scene.}
     \label{fig:nsademo}
 \end{figure*}
 
\subsection{Current Scenario}
\label{sec:cursec}
CNNs, like many other neural network architectures, are black-box, which means their working cannot be understood entirely. The situation became further concerning when Szegedy et al. \cite{szegedy2013intriguing} highlighted neural networks' brittleness against corruption. Several demonstrations highlighted the ability of small (even imperceptible) noise in the image (popularly known as adversarial attacks) to cripple perfectly trained models \cite{szegedy2013intriguing}, \cite{eykholt2018robust}, \cite{athalye2018synthesizing}. In most cases, the addition of noise required precise manipulation of the image's pixel values digitally (FGSM\cite{goodfellow2014explaining}, PGD \cite{madry2017towards}, C\&W \cite{carlini2017towards}), limiting its applicability. To translate adversarial attacks to practical scenarios, Brown et al. ensured the printability of attack and proposed Adversarial Patches \cite{brown2017adversarial}, \cite{10.1145/3579988.3585054}. 

The numerous proposals of varied corruption in recent years necessitate robust model training. Eventually, the concern over the inconsistent behavior gave rise to several defense methodologies \cite{akhtar2018threat}, \cite{sharma2022adversarial}, \cite{icaart22}. A defense can be broadly classified as either model agnostic (e.g., using saliency map) \cite{chou2020sentinet}, \cite{xiang2022patchcleanser} or model dependent (adversarial training) \cite{madry2017towards} where the network weights are learned to tackle the corruption. Typically, the design of a defense either requires access to model parameters or the attack should be localized and perceptible to be detectable. The arms race between robust defenses and newer malicious attacks overcoming them is an ongoing challenge.

\subsection{Motivation}
\label{sec:moti}
The robustness of an AI model is more than just ensuring performance against adversarial noises. With the advent of mathematically formulated  attacks, the focus of the literature has shifted away from the fundamental natural corruptions (e.g., snow, dust, rain). The chances of such natural disturbances occurring in the real world are far more likely than adversarial attacks. We acknowledge that the potency of natural corruption is lower than that of adversarial attacks, yet designing a robust model requires their consideration. Most commonly, researchers have been using image augmentations \cite{shorten2019survey} to train resilient CNN models \cite{rebuffi2021data}, \cite{rebuffi2021fixing} against natural noises. In image augmentation, corruptions in the form of transformations or disturbances are applied to the training images. In essence, the network parameters are learned to make robust predictions over corrupted images \cite{madry2017towards}. However, in practice, the accessibility to model parameters is sometimes infeasible, and model training with augmented images is expensive, limiting the technique's applicability.

 Inspired by adversarial patch \cite{brown2017adversarial}, Salman et al. proposed unadversarial example \cite{salman2021unadversarial} to tackle natural corruptions. A normal image can be easily transformed into an unadversarial one by adding an inverse adversarially trained patch to it. In this technique, the loss is backpropagated to learn features/patterns in the scene rather than network weights, as in the case of adversarial training.  The learned patterns are either restricted to a localized region as a patch or are printed over the target object's body. Designing unadversarial examples does not require access to model parameters but can still improve the prediction confidence score. 
 Building over the idea of Salman et al., the authors in \cite{wang2022defensive} proposed a more vigorous defense to achieve better robustness. Additionally, they demonstrated the effect of unadversarial examples on the distribution shift and utilized class-level information for better performance. A similar work called collaborative adversarial training (CAT) \cite{li2022collaborative} demonstrates a new distance metric for generating unadversarial examples for adversarial robustness. 
 
 The existing works on unadversarial examples institute the idea of implicit  scene robustness without relying on model training. However, they all contain unnatural trained patterns in the scene. In the unadversarial examples proposed in \cite{salman2021unadversarial}, the trained patterns over the target object look suspicious and make it easily noticeable. Similarly, in \cite{wang2022defensive}, an irregular, thick boundary-trained patch is made around the image, which looks unnatural. Making a boundary for any image is only feasible after capturing it through the camera. Although the vision behind unadversarial examples is highly appreciable, there are better ways to generate them than the existing techniques.
 
\subsection{Our Contribution}
Our work draws significant inspiration from the unadversarial example design technique. As per the discussion in Section \ref{sec:moti}, the primary goal should be to make the patches added to unadversarial images look natural. In the context of our work, we call these patches artifacts. The `artifacts' can be defined as artificially generated objects that initially do not exist in the scene, but are placed intentionally. This work proposes an artifact training framework to design naturalistic support artifacts (NSAs). The location of artifacts can vary depending on the context of the image. Figure \ref{fig:nsademo} demonstrates how the artifacts will look in the scene and their ability to boost the prediction confidence score against natural corruption.  

Similar to unadversarial examples, the NSA does not require knowledge of model architecture or accessibility to its parameters. Hence, robust prediction is not a result of the model itself but rather due to the placement of NSA in the scene. Here, we assume that the NSA designer has physical and digital access to the scene to train and place the artifacts in the scene. This is a fair assumption in most cases as the model designer often also has the accessibility to the scene or at least has control over the inputs to the model \cite{salman2021unadversarial}. Additionally, the NSAs have excellent visual fidelity in the scene. Since the artifacts help the model's prediction, we use the term `support' in NSA. The characteristics of the NSA are summarised below:

\begin{itemize}
    \item \textbf{High Visual Fidelity: }The NSAs placed in the scene look natural without any suspicious patterns. 
    \item \textbf{Universal Training: }The NSAs can be universally trained for all physical corruptions. However, fine-tuning does improve the prediction robustness.
    \item \textbf{Model Agnostic: }The framework of NSAs training is independent of classifier and generator architecture.
    \item \textbf{Scalable: }The number of NSAs in the scene are scalable by increasing varied pre-trained artifact Generators.
\end{itemize}

\section{Background}

This section discusses a generic model and generator formulation concerning naturalistic support artifacts. 

\subsection{NSA Formulation}

\subsubsection{Model Formulation}
\label{sec: modelform}
Assume we have an input RGB image $\bm{x} \in \mathcal{X}$, where $\mathcal{X} \in {\mathbb{R}}^{w\times h \times c}$. The 
$w$, $h$, and $c$ represent the width, height, and color channels in the image, respectively. The image is normalized in the range $[0,1]$ to ensure printability. The CNN model $\mathcal{F}: \mathcal{X} \rightarrow \mathcal{Y} $ produces the probabilistic output vector $\vec{y} \in \mathcal{Y}$, where $\mathcal{Y} \in {\mathbb{R}}^{n}$ and $n$ is the total number of classes present in the dataset. Each element of $\vec{y}$ is the probability of classifying image $\bm{x}$ to a corresponding class in the dataset. Deriving from the classification probability, the confidence score can be formulated as the prediction probability $\times$ 100. This score signifies the confidence of a model in the specific prediction. The class corresponding to the highest probability in $\vec{y}$ is the model's predicted class $k$ of an image $\bm{x}$. The $k$ can vary from $\{0,1,....,n-1\}$. The equation formulating the model behavior is given as

\begin{equation}
    k = \operatorname*{argmax} [\mathcal{F} (\vec{y} |\bm{x})]
    \,,\label{eq:model}
\end{equation}

where the model $\mathcal{F}(\vec{y}|\bm{x})$ outputs the probability vector $\vec{y}$ for a given image input $\bm{x} \in \mathcal{X}$. 

\begin{figure*}[t]
     \centering 
     \includegraphics[scale=0.25]{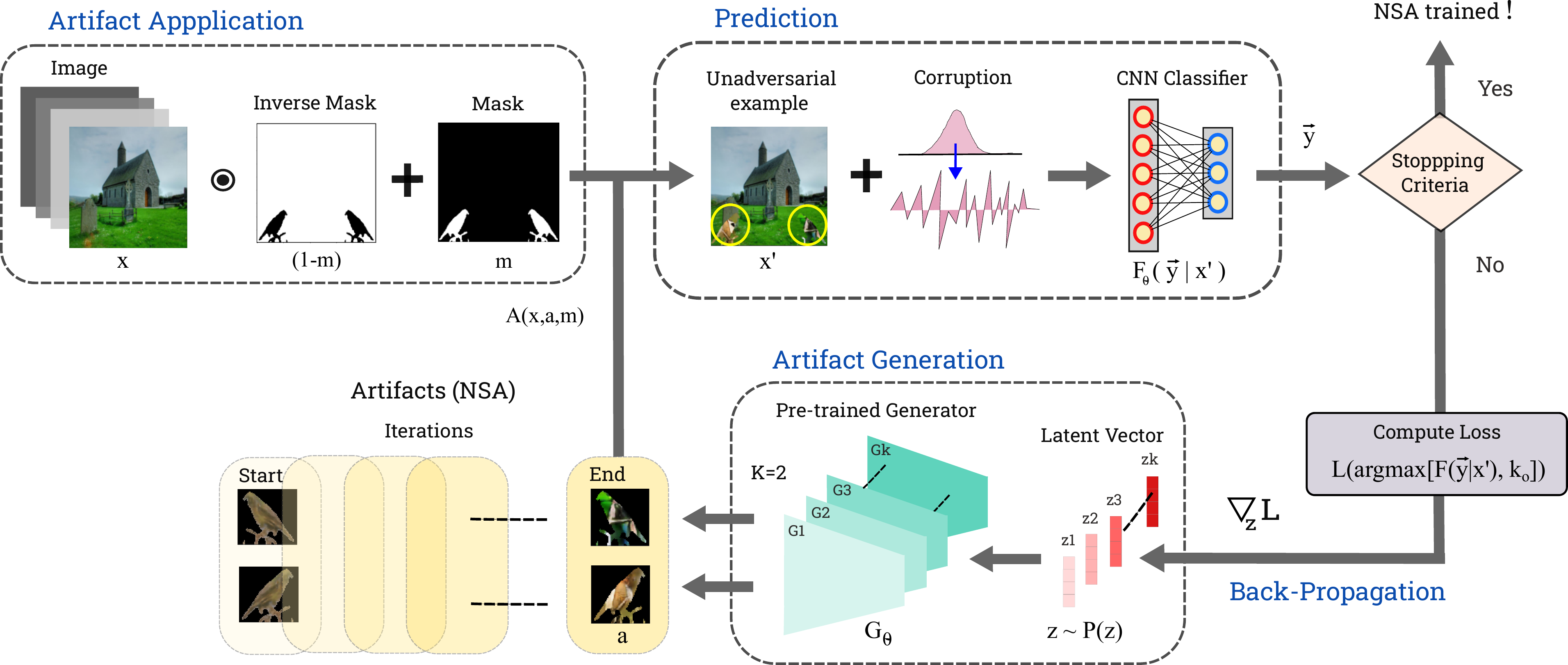}
     \caption{
     Artifact training framework for designing NSAs. The unadversarial image is formed using two artifacts for illustration. Note that the $\odot$ represents Hadamard product. The yellow circle highlights the presence of NSAs.}
     \label{fig:trainloop}
 \end{figure*}

\subsubsection{Generator Formulation} The Generator $\mathcal{G} : {\mathbb{R}}^{n} \rightarrow {\mathbb{R}}^{\bar{w}\times \bar{h} \times c}$ is used to create naturalistic artifact $\art = \mathcal{G}(z)$, $\art \in {\mathbb{R}}^{\bar{w}\times \bar{h} \times c}$, where $\bar{w}$ and $\bar{h}$ represent width and height of the artifact. The channels $c$ of the artifact are the same as that of the image. The $z$ is the input latent vector randomly sampled from the noise distribution $P(\mathbf{z}) \in {\mathbb{R}}^{n}$.  The framework of designing unadversarial examples is similar to that of \cite{salman2021unadversarial}, but the generator's inclusion in the training loop distinguishes our method. A Deep Convolutional Generative Adversarial Network (DC-GAN) is trained in unsupervised fashion with the Wasserstein GAN with gradient penalty (WGAN-GP) loss as shown in Equation \ref{eq:wgangp}. 

\begin{equation}
\begin{aligned}
    \min_{\mathcal{G}} \max_{\mathcal{D}} \hspace{0.2cm} \mathbb{E}_{\tilde{x} \sim P_g} \left[ \mathcal{D}(\tilde{x})\right] - \mathbb{E}_{x \sim P_r} \left[ \mathcal{D}(x)\right] \\ 
    + \lambda \hspace{0.1cm} \mathbb{E}_{\hat{x} \sim P_{\hat{x}}} \left[ {(||\nabla_x \mathcal{D}(\hat{x})||_2-1)}^2 \right] 
    \,,\label{eq:wgangp}
\end{aligned}
\end{equation}

where $x$ is the real image and $\mathbb{P}_r$ is the distribution of real images. The $\tilde{x}$ is the fake image ($\tilde{x} = \mathcal{G}(z)$) from the distribution $\mathbb{P}_g$ of generated images. The $\mathbb{P}_{\hat{x}}$ is the distribution representing intersection between real and fake images. The min-max optimization ensures that the generated images look similar to real ones. Authors in \cite{gulrajani2017improved} have shown that WGAN-GP ensures stable training with good visual fidelity among generated images. The dataset for training DC-GAN depends on the type of artifact to be placed in the scene, like birds, ball etc. The normally trained generator is utilized as a NSA generator during the artifact training.

\subsubsection{Artifact Formulation}
\label{sec: attackform}
We formulate artifact creation as an optimization problem. The goal is to ensure that the artifacts help minimize the loss of classifying the image to the original label $k_o$, given as $\mathcal{L}(\operatorname*{argmax} [\mathcal{F} (\vec{y} |\bm{x}')], k_o)$. The gradient information of the prediction loss is backpropagated to update the artifacts' pixel values iteratively. We design a background removal technique to create a mask for applying artifacts in the scene, and we discuss it in detail in Section \ref{sec: atkproc}. The unadversarial example $x'$ formed after placing artifact $ \art$ on the image $x$ is shown in equation \ref{eq:attack}

\begin{equation}
    \bm{x}' = (1-m)\odot\bm{x} + m \odot \art
    \,,\label{eq:attack}
\end{equation}
where $\bm{x}' \in \mathcal{X}, \art \in {[0,1]}^{\bar{w} \times \bar{h} \times c}$ is the artifact and $m \in M \subset {\{0,1\}}^{w \times h \times c}$ represents the binary pixel block to mask. The mask specifies the patch's area and location over the image. The $\odot$ is the Hadamard operator which denotes the element-wise multiplication of pixels between two matrices. Additionally, artifact' pixel values are clipped at every iteration to stay within the valid RGB range to ensure the printability.

\section{Artifact Training}
\label{sec: atkproc}
The design of prediction-supporting artifacts has been inspired from \cite{salman2021unadversarial}, where the authors show that an adversarial attack can be turned into an unadversarial example using slightly modified loss function as shown in Equation \ref{eq:unadv}. 

\begin{equation}
\label{eq:unadv}
\begin{aligned}
    \delta_{adv} = \argmax_{\delta \in \Delta}{\mathcal{L}(\mathcal{F}(x+\delta),y)} \\
    \delta_{unadv} = \argmin_{\delta \in \Delta}{\mathcal{L}(\mathcal{F}(x+\delta),y)}
\end{aligned}
\end{equation}
where $y$ is the original label and $\delta$ is the perturbation bounded by $\Delta$, added to the image $x$ to form an unadversarial example. In addition, the idea of generating natural objects has been extended from the framework proposed in \cite{doan2022tnt}. 

An illustration of the artifact training framework has been shown in Figure \ref{fig:trainloop}. The framework consists of three components: artifact generation, artifact application, and prediction, which are explained in detail as follows:

\vspace{0.1cm}
\textbf{Artifact Generation: } The artifact generation procedure requires a pre-trained generator to generate artifacts from a given distribution. The parameters of the generator are fixed during the training. The input to the generator is an $n$-dimensional latent vector $z \in \mathbb{R}^{n}$. To boost the prediction score, we propose using multiple artifacts in the scene. The advantage of using multiple artifacts is three-fold: First, it increases the number of pixels that can be manipulated. With more artifacts, we have higher control in the scene. Second, placing multiple artifacts is better as a single artifact could become unnaturally large while trying to achieve satisfactory performance. Lastly, typically there are seldom any physical limitations to placing more than one artifact in the scene. Additionally, multiple artifacts also helps in maintaining visual fidelity as we have more freedom to place artifacts at different locations in the scene. 

\begin{algorithm}[t]
\label{alg:1}
\caption{Artifact Training Procedure}
\label{alg:1}
\KwIn{\small Image $\bm{x}$, Generator $\mathcal{G}$(.), Classifier $\mathcal{F}$(.), corruption function $N$(.),  corruption severity $s$, target label $y_t$, loss function $\mathcal{L}$, target confidence score $p_t$, old confidence score  $p_o$, new confidence score $p_n$, convergence tolerance $c_t$, number of artifacts $k$}
\BlankLine%
\For{i=$1, \cdots, k$}{
$\mathbf{z_i} \sim P(\mathbf{z})$ \hspace{0.1cm} // \texttt{\scriptsize sample a random latent vector} 

$\art^{n}_i = \mathcal{G}(z)$ \hspace{0.1cm} // \texttt{\scriptsize generate naive artifact}

$\bar{\art}_i = \texttt{\small bg\_remove}(\art^{n}_i)$ \hspace{0.1cm} // \texttt{\scriptsize remove background}

$\mask= \opA(\bar{\art}_i)$ \hspace{0.1cm} // \texttt{\scriptsize design mask with patch applicator}
}
\While{$\mathcal{F}(y=y_t|\bm{x}') < p$ }
{
    \For{i=$1, \cdots, k$}{
      $\art^{t}_i = \mathcal{G}(z_i)$ \hspace{0.1cm} // \texttt{\scriptsize generate naive artifact}
      
      $\bm{x}'= (1-\mask)\odot\bm{x}+\mask\odot \art^{t}_i$ \hspace{0.1cm} // \texttt{\scriptsize create unadversarial image}
      
      $p_o = \max[\mathcal{F}(y=y_t|\bm{x}')]$
      
      $\vec{y}= \mathcal{F}(\bm{x})$
      
      $\caLt= \ell(\vec{y},y_t)= \mathcal{F}(y=y_t|\bm{x}')$ \hspace{0.1cm} // \texttt{\scriptsize Loss function for the targeted attack}
      
      $\naDt= \frac{\partial}{\partial \bm{z}} \caLt$ \hspace{0.1cm} // \texttt{\scriptsize Calculating gradient w.r.t image pixels}
      
      $z_i = z_i - \epsilon\cdot  \naDt$ \hspace{0.1cm} // \texttt{\scriptsize Update pixels to minimise targeted loss}
      
      $i= i+1$ 
}
    $ \bm{x}' =  N(\bm{x}', s)$ \hspace{0.1cm} // \texttt{\scriptsize apply corruption}

      $p_n = \max[\mathcal{F}(y=y_t|\bm{x}')]$ \hspace{0.1cm} // \texttt{\scriptsize confidence prediction}
      
      \If{$\|p_n - p_o\| \leqslant c_t$}{
      
      $\textbf{break}$ 
      }
}
\end{algorithm}

As shown in Figure \ref{fig:trainloop}, our framework facilitates simultaneous training of all artifacts without additional complexity. It is necessary as we want all the artifacts to work in tandem and complement each other. Typically, to improve the prediction robustness, the artifacts learn patterns by focusing on salient features. Hence, if we artifacts train individually, they will have similar patterns (primarily drawn from the salient region). Any abnormality in the non-salient regions has the potential to hamper performance. With simultaneous training, the artifact tends to support each other and focus on complementary features. If one artifact focuses on the salient object, the other tries to derive information from the rest of the scene. Hence, infusing more varied scene context into the artifacts is possible, ultimately increasing the chances of generating better artifacts. Also, the type of artifact can vary as we can use different generators producing different types of artifacts (for example, generator $\mathcal{G}1$ for \emph{bird}, generator $\mathcal{G}2$ for \emph{ball}). It increases the freedom of creating an unadversarial example. 

\vspace{0.1cm}
\textbf{Artifact Application: }Placing the artifacts requires one to place a suitable mask $m$  based on the context of the image. The mask is designed in two stages: First, we determine a custom threshold value for each artifact by trial and error to remove the background. This process converts the artifact's image into a binary image (sub-mask). The size of each sub-mask will be $\bar{w} \times \bar{h}$ (same as that of generated artifact from $\mathcal{G}$), where the background will be black (pixel value 0) and artifact body will be white (pixel value 1) due to thresholding. Next, we place the sub-masks corresponding to each artifact on a black (pixel value 0) background of ($w \times h$), the same as the original image. The sub-masks are placed using a patch application operator $\opA$ to apply artifacts at required locations and with relevant orientations. For the purpose of demonstration, we present two bird artifact based mask in Figure \ref{fig:trainloop}. The mask is kept static during iterations because it needs to be pre-designed as per the scene. However, it is not mandatory and can be changed dynamically if the application demands it.  

\textbf{Prediction: }Once all the artifacts are placed in the scene, the unadversarial example is ready for prediction. To achieve higher robustness, we apply the corruption of interest on the image (refer Figure \ref{fig:trainloop}). In this way the artifacts can be fine-tuned against specific corruptions for better performance. The image is sent to a CNN model, for which we need to boost the prediction confidence. The model needs to be pre-trained on the dataset of interest. Note that even though we help in the CNN's prediction, the model's parameters are fixed. The model behaves as a black-box in the framework and takes an unadversarial example as input and predicts the output.

 \vspace{0.1cm}
The overall artifact training procedure can be summarised as follows: First, a loss function is decided as per the application. Like a typical neural network training, the loss gradient over the output prediction is backpropagated. While creating an unadversarial example, we calculate the loss gradient with respect to the artifact's pixels rather than network weights. However, it leads to unrestricted modification of the artifact's pixel values, eventually leading to suspicious and unnatural patterns. Also, these patterns are in the form of a geometric shape, which may not be relevant in the context of the scene. To avoid unnatural patterns and learn meaningful objects, we utilize a generator in the training loop to include the additional constraint of producing naturalistic artifacts. Hence in this work, rather than directly updating the artifact's pixels, we update the generator's input latent vector. If we apply multiple artifacts, all the latent vectors for respective generators are updated simultaneously. The generator's output is the desired set of artifacts/NSAs. These NSAs are then applied over the target image using a suitable mask. The corrupted unadversarial image is then sent to the model for prediction. The training is continued until a target prediction confidence score is achieved. We also include additional stopping criteria to stop the training if there is an insufficient improvement of the prediction score for a set of subsequent iterations. The detailed procedure is explained in Algorithm \ref{alg:1}.

\section{Experiments and Results}
\label{sec:expandres}

In this section, we discuss the results of a set of experiments to inspect the performance of NSA.
\subsection{Experimental setup}
We use two different models to validate our attack: VGG16 \cite{simonyan2014very}, and ResNet18 \cite{he2016deep}. They have distinct backbone architectures, increasing the possibility of diversity in results. 
These models are trained on Imagenette, a simple and commonly used benchmark dataset with 10,000 images shared almost equally between 10 classes. Each image is 224$\times$224 pixels. Imagenette is derived as a subset from the commonly used benchmark dataset: ImageNet \cite{krizhevsky2017imagenet}. 

For training the artifact generator, we used PyTorch's \texttt{TorchGAN} \cite{Pal2021} library for ease of implementation. We decided on birds as an artifact for our applications and used a custom subset of the Bird-400 dataset for the training. We found 300 epochs to be sufficient to train the artifact generator with WGAN-GP loss. We decided $n$ = 128 as the dimension for input latent vector to the generator. We found that lower dimensions ($n$ = 64) generated low quality artifact and we did not observe any additional improvement with higher dimensions ($n$ = 256). The output of the generator is an 64$\times$64$\times$3 artifact.

With the \texttt{Cross Entropy} loss, we used PyTorch's inbuilt \texttt{Adam} optimizer with a learning rate of 0.1 for the artifact training. \texttt{OpenCV}'s image thresholding function is utilized to remove the background of generated images from the artifact generator to design the mask. Overall, the NSA's cover around 5\% of the total image area. The corruptions are introduced in the scene using an off-the-self Python library \texttt{imagecorruptions} \cite{michaelis2019dragon}. All the experiments were carried out on the single Nvidia RTX A6000. 

\begin{figure}[t]
\begin{center}
     \includegraphics[width = 8cm, height = 10cm]{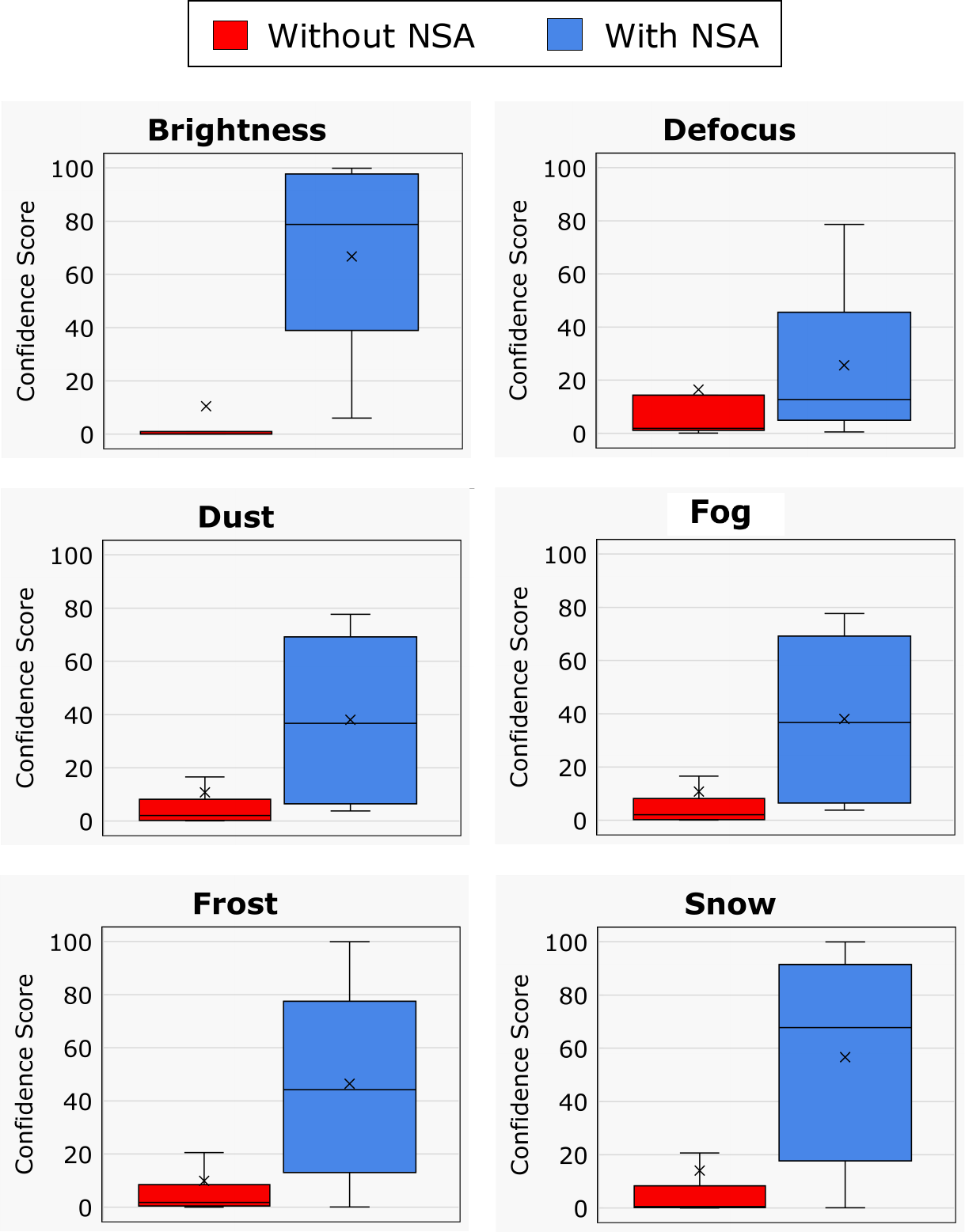}
  \caption{Comparison of prediction confidence score against corruption distributions with and without NSA. The '$\times$' symbol in the plot represents the mean. The bottom edge of box represents the 25th percentile and the upper edge represents the 75th percentile, with the middle line inside the box being the median of the data. All the corruptions belong to Level 3.} 
  \label{fig:conf_score}
\end{center}
\end{figure}

\begin{figure*}[t]
\centering
\begin{subfigure}{0.32\textwidth}
  \centering
  \includegraphics[width = 5.3cm, height = 3.1cm]{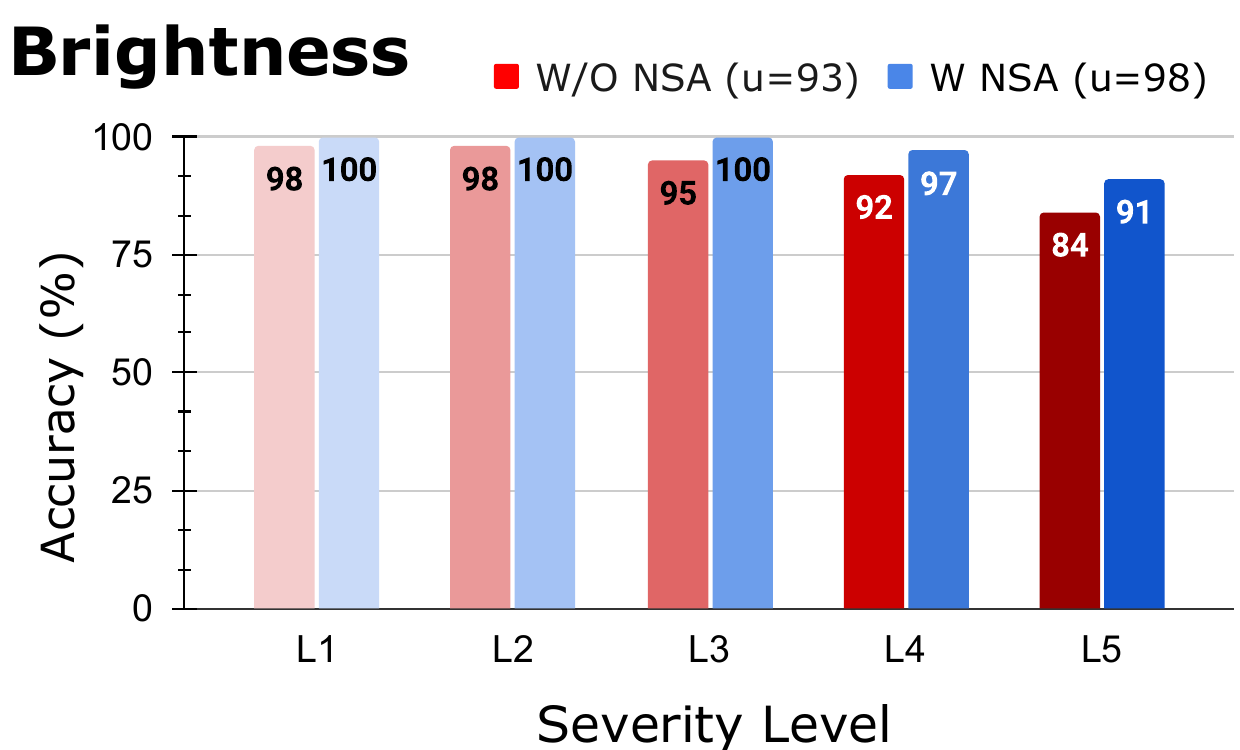}  
  \phantomsubcaption{}
  \label{fig:r_B}
\end{subfigure}
\begin{subfigure}{0.32\textwidth}
  \centering
  \includegraphics[width = 5.3cm, height = 3.1cm]{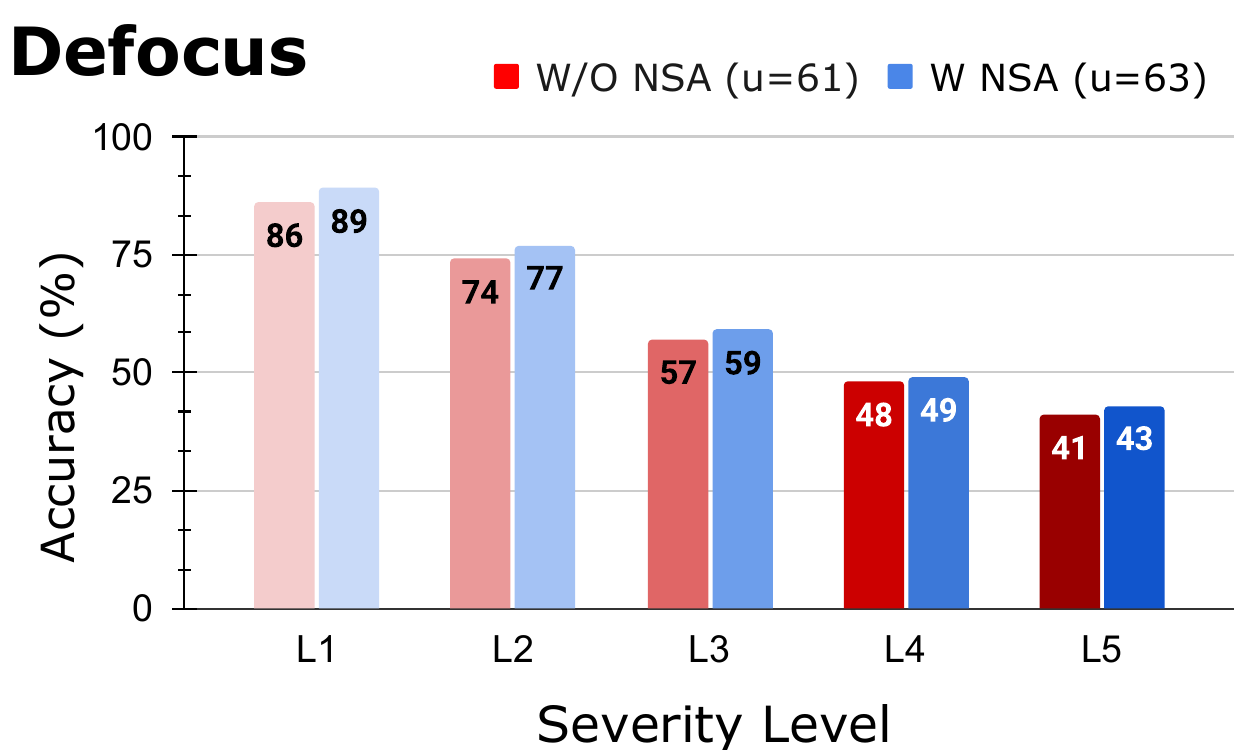}  
  \phantomsubcaption{}
  \label{fig:r_Df}
\end{subfigure}
\begin{subfigure}{0.32\textwidth}
  \centering
  \includegraphics[width = 5.3cm, height = 3.1cm]{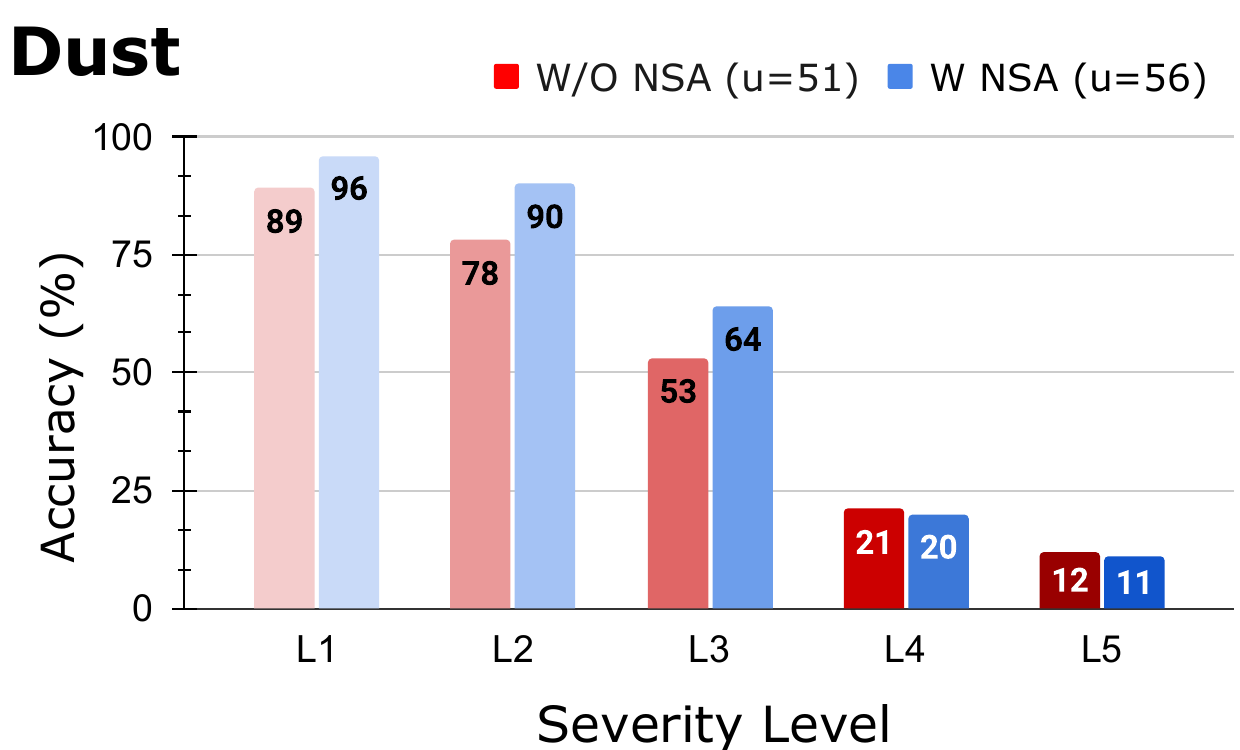}  
  \phantomsubcaption{}
  \label{fig:r_D}
\end{subfigure}

\begin{subfigure}{0.32\textwidth}
  \centering
  \includegraphics[width = 5.3cm, height = 3.1cm]{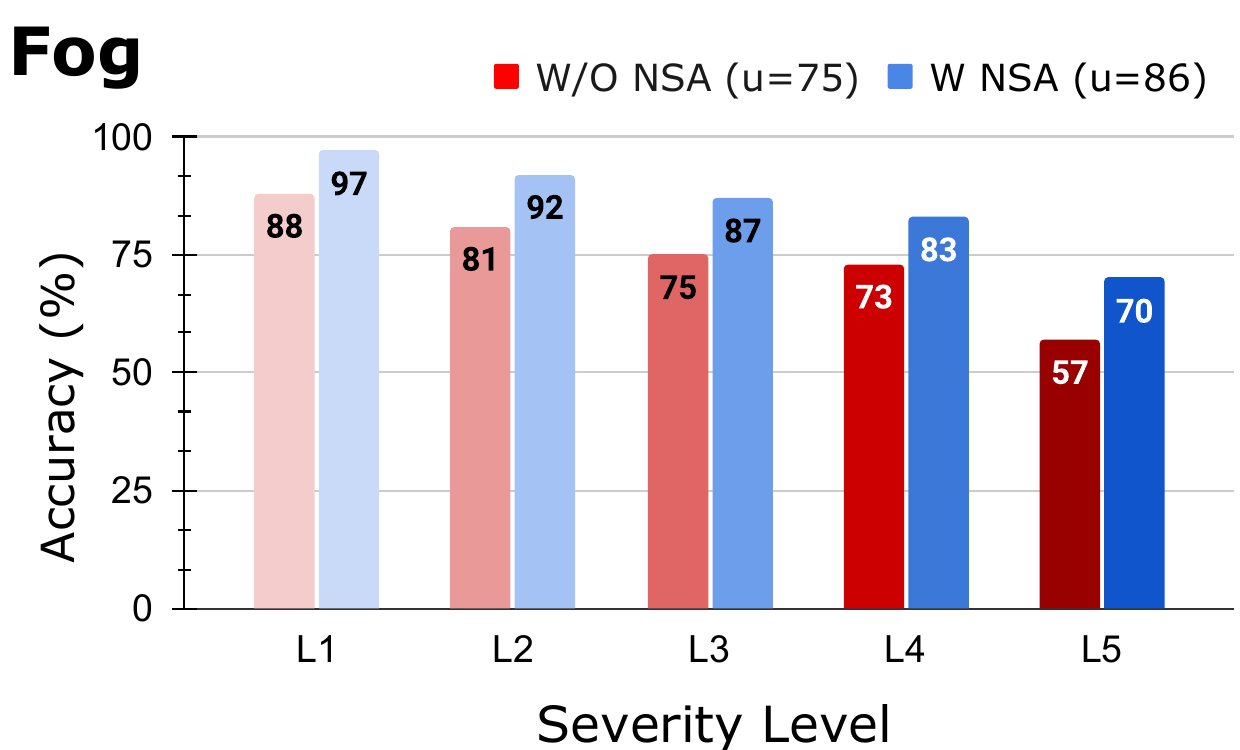}  
  \phantomsubcaption{}
  \label{fig:r_Fg}
\end{subfigure}
\begin{subfigure}{0.32\textwidth}
  \centering
  \includegraphics[width = 5.3cm, height = 3.1cm]{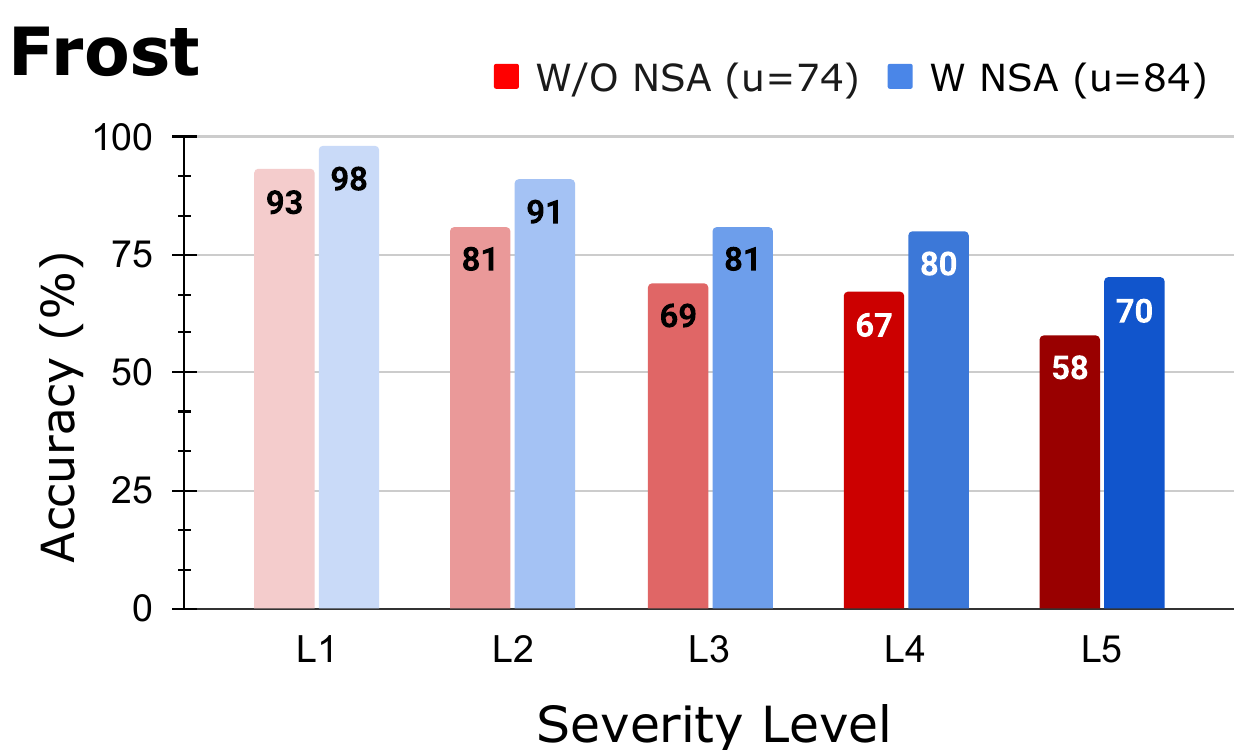}  
  \phantomsubcaption{}
  \label{fig:r_Fr}
\end{subfigure}
\begin{subfigure}{0.32\textwidth}
  \centering
  \includegraphics[width = 5.3cm, height = 3.1cm]{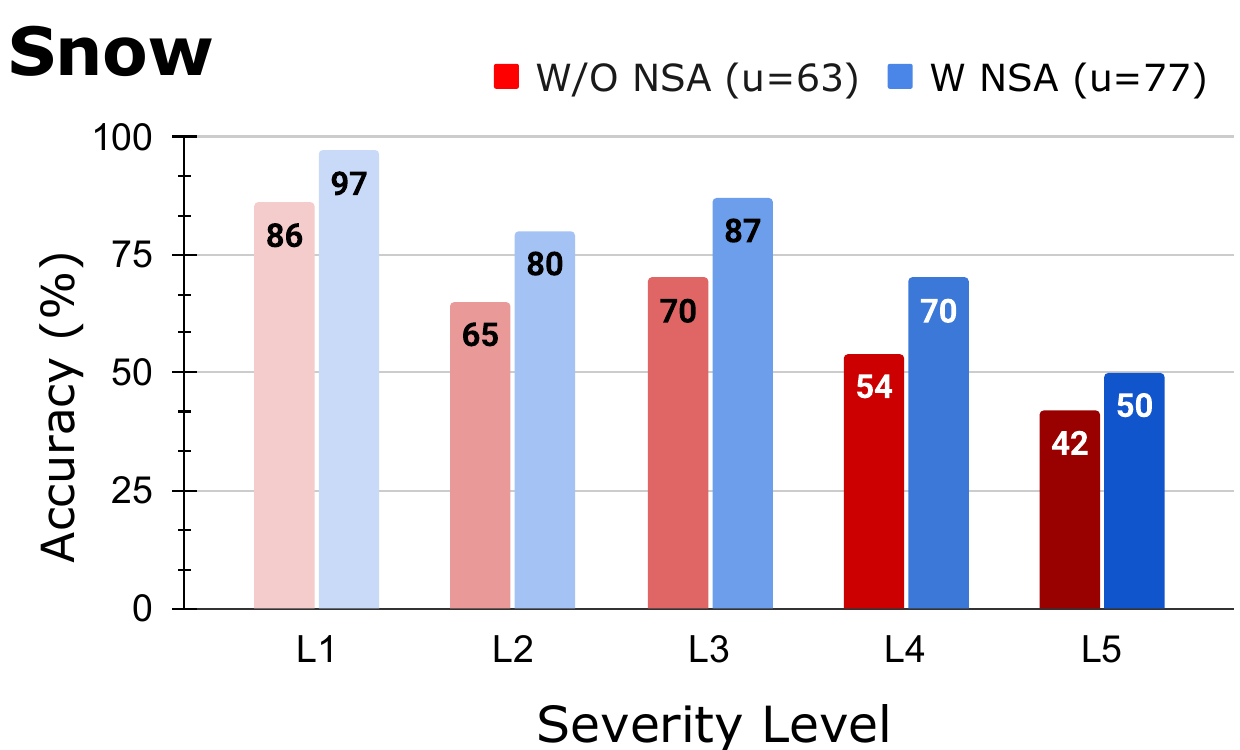}  
  \phantomsubcaption{}
  \label{fig:r_S}
\end{subfigure}

\caption{Analysis of naturalistic support artifacts (NSA) against physical corruption for ResNet18 architecture. 
}
\label{fig:resnet}
\end{figure*} 

\begin{figure*}[t]
\centering
\begin{subfigure}{0.32\textwidth}
  \centering
  \includegraphics[width = 5.3cm, height = 3.1cm]{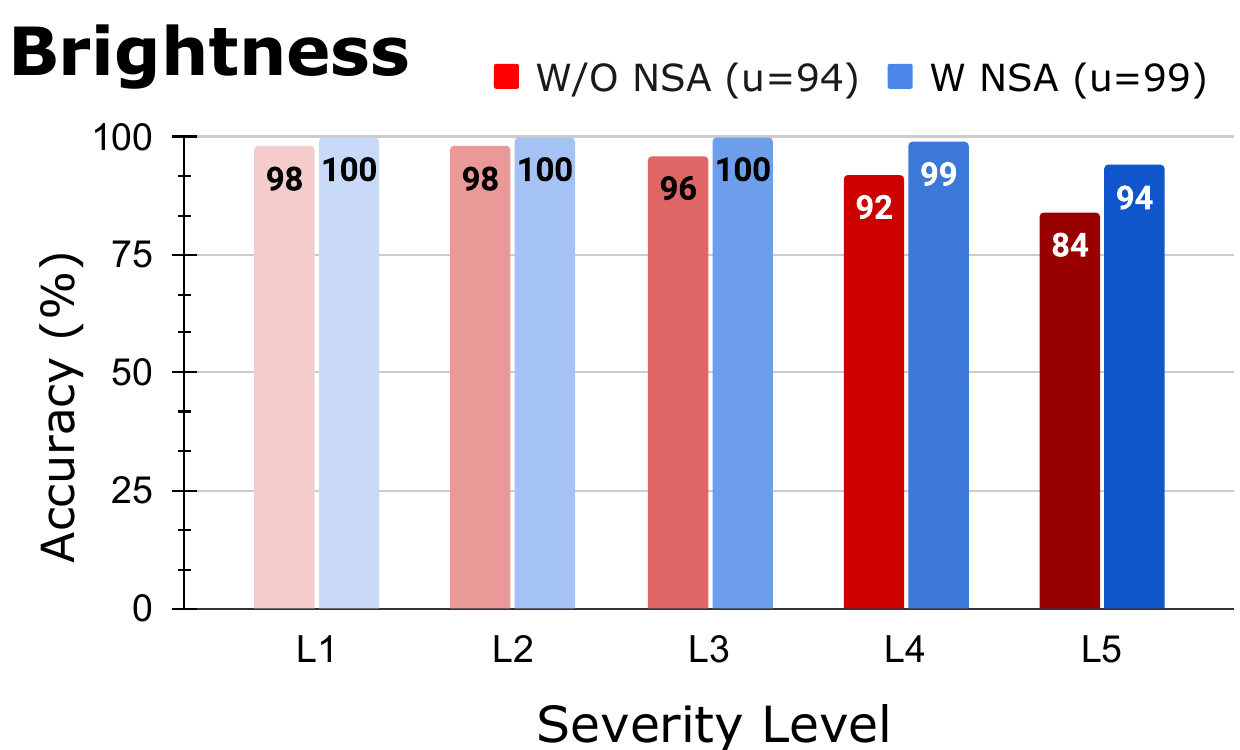}  
  \phantomsubcaption{}
  \label{fig:r_B}
\end{subfigure}
\begin{subfigure}{0.32\textwidth}
  \centering
  \includegraphics[width = 5.3cm, height = 3.1cm]{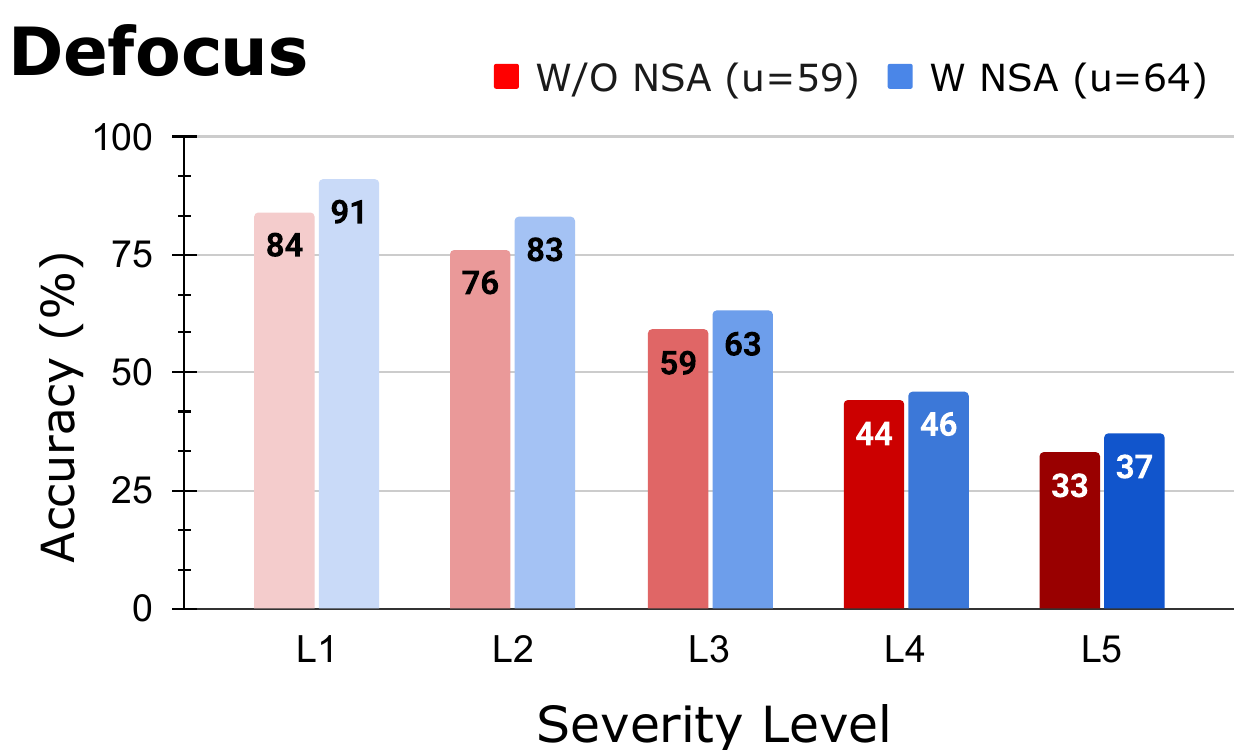}  
  \phantomsubcaption{}
  \label{fig:r_Df}
\end{subfigure}
\begin{subfigure}{0.32\textwidth}
  \centering
  \includegraphics[width = 5.3cm, height = 3.1cm]{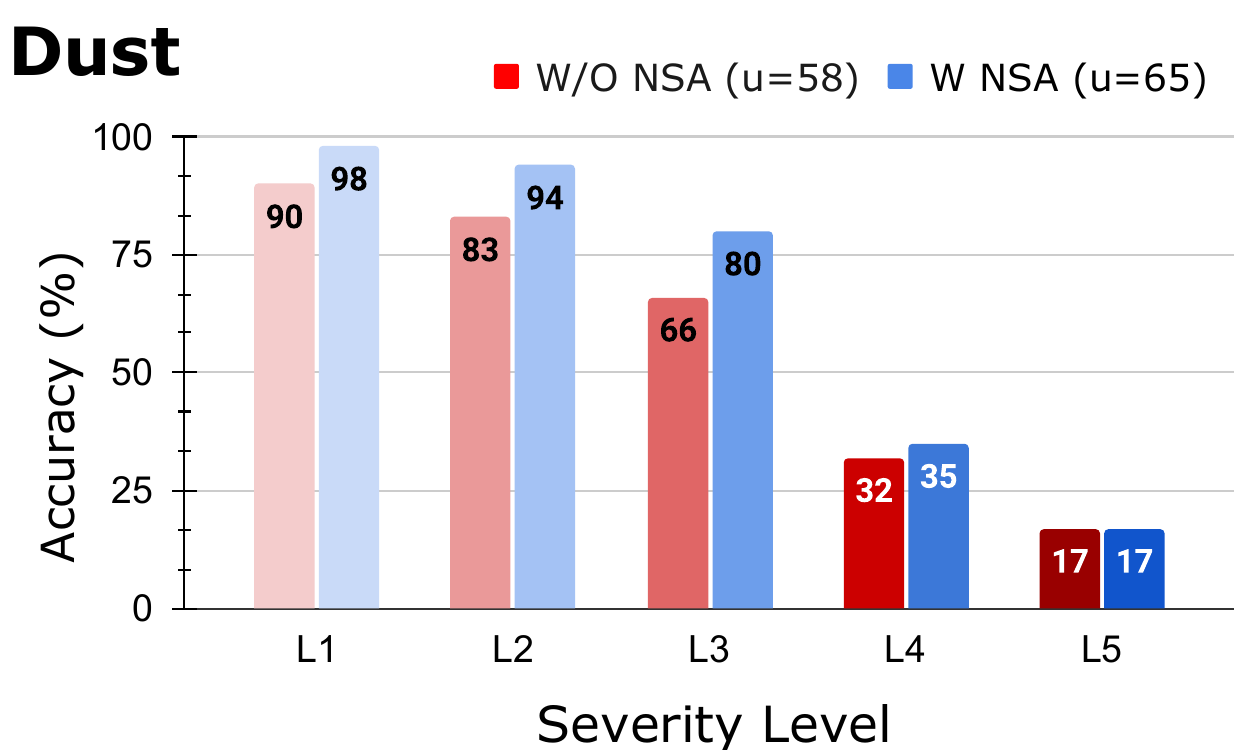}  
  \phantomsubcaption{}
  \label{fig:r_D}
\end{subfigure}

\begin{subfigure}{0.32\textwidth}
  \centering
  \includegraphics[width = 5.3cm, height = 3.1cm]{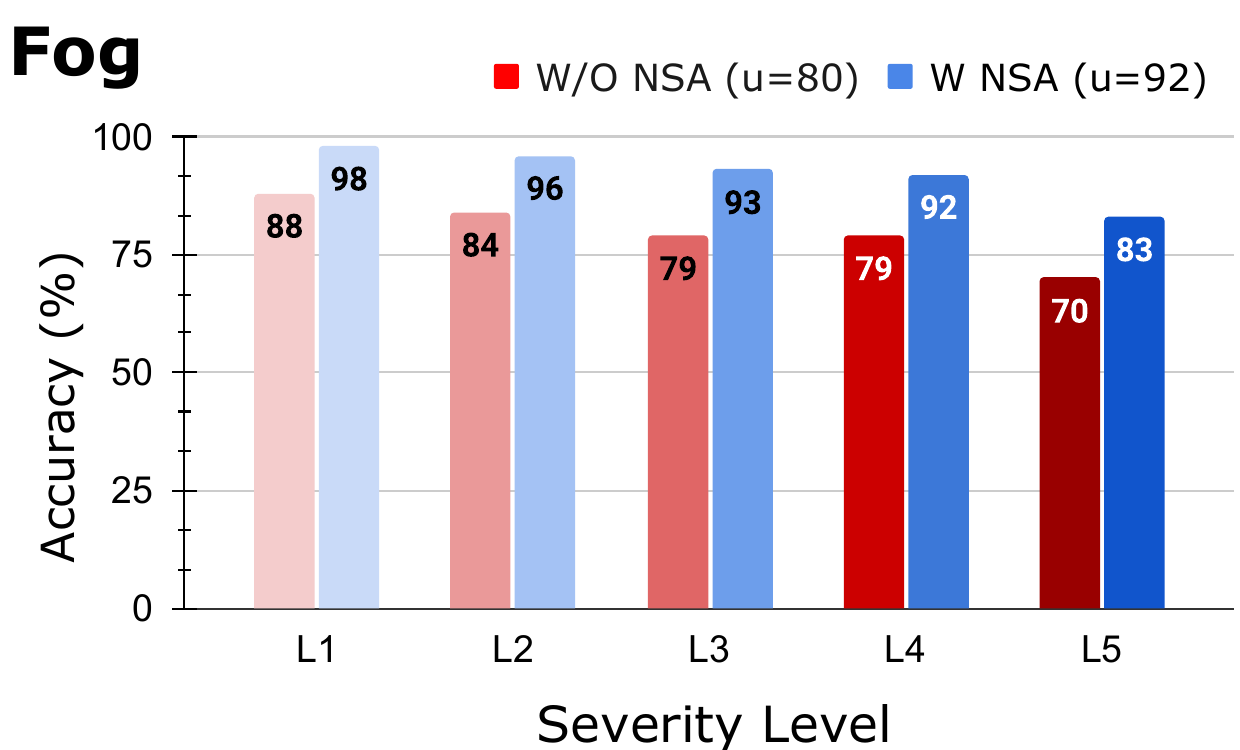}  
  \phantomsubcaption{}
  \label{fig:r_Fg}
\end{subfigure}
\begin{subfigure}{0.32\textwidth}
  \centering
  \includegraphics[width = 5.3cm, height = 3.1cm]{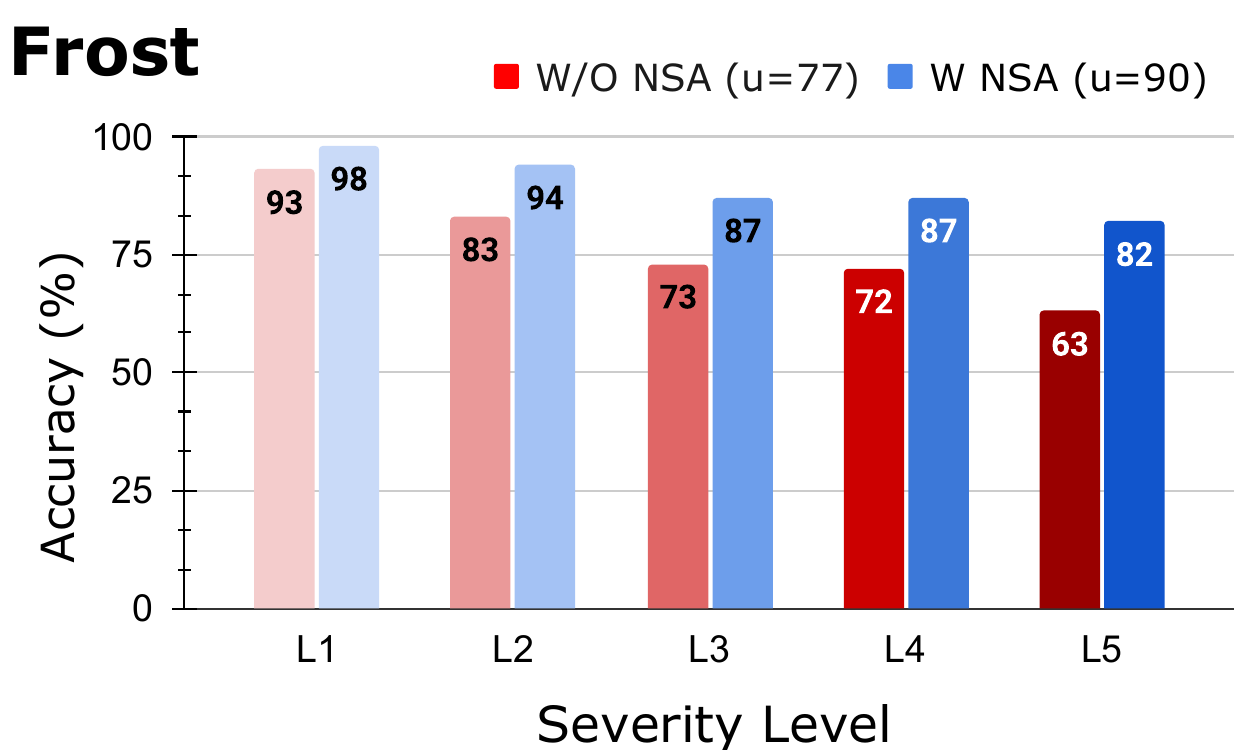}  
  \phantomsubcaption{}
  \label{fig:r_Fr}
\end{subfigure}
\begin{subfigure}{0.32\textwidth}
  \centering
  \includegraphics[width = 5.3cm, height = 3.1cm]{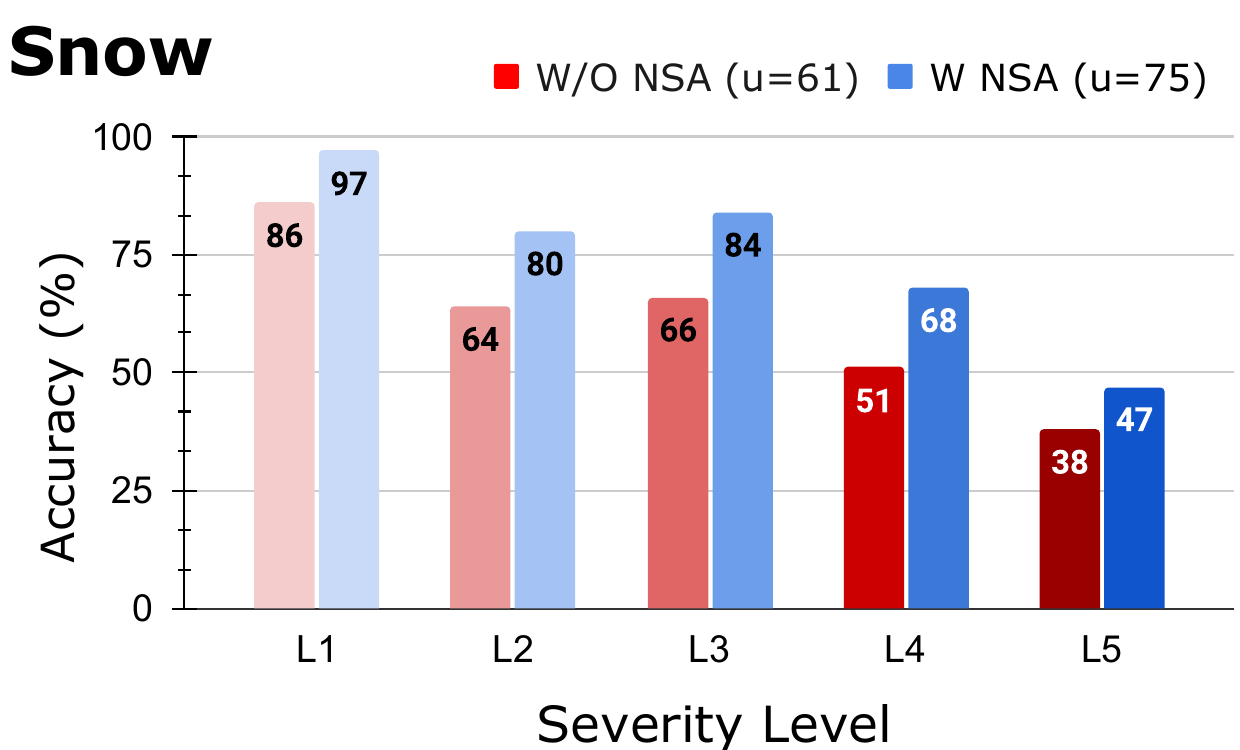}  
  \phantomsubcaption{}
  \label{fig:r_S}
\end{subfigure}

\caption{Analysis of naturalistic support artifacts (NSA) against physical corruption for VGG16 architecture. 
}
\label{fig:vgg}
\end{figure*} 

\subsection{Experiment I: Analysis of NSA's impact on the prediction confidence score}
As a part of this experiment, we analyze how NSAs increase confidence in the prediction. Since all the natural corruption derives from an underlying distribution, we record confidence scores based on 1000 noise samples from a specific corruption distribution, applied to a random image from each of the ten classes. The score is calculated from the output classification probability vector of the CNN model multiplied by 100. Hence, each plot consists of confidence score values from the prediction on 10,000 adversarial images for a specific corruption, which is sufficient to evaluate the impact of the NSAs. Additionally, the values are averaged across the ResNet18 and VGG16 model.

\makeatletter
\newcommand{\thickhline}{%
    \noalign {\ifnum 0=`}\fi \hrule height 1pt
    \futurelet \reserved@a \@xhline
}
\makeatletter
\newcommand{\midhline}{%
    \noalign {\ifnum 0=`}\fi \hrule height 0.7pt
    \futurelet \reserved@a \@xhline
}
\newcolumntype{'}{@{\vrule width 0.1pt}}
\newcolumntype{"}{@{\vrule width 0.9pt}}
\makeatother

In Figure \ref{fig:conf_score}, we observe that the median and mean confidence scores for NSA-applied images are higher than the ones without them for all corruption types. We notice that for brightness, fog, frost, and snow, the 75\% percentile score without NSA is lower than that of 25\% percentile with NSA. Overall, we observe about 4 times improvement in the mean confidence score with NSA across all corruptions. The highest impact is against brightness (6.4 times), and the lowest is for defocus (1.6 times). Overall, the NSA is shown to considerably improve the confidence, which eventually also improves the adversarial accuracy as demonstrated in the Experiment I.  

\subsection{Experiment II: Analysis of NSA's impact on the adversarial accuracy}
We evaluated the ability of the NSA to improve the adversarial accuracy of ResNet18 and VGG16 classifiers against physical corruption. We used the same test set for both models to maintain consistency. We chose six prominent corruptions that frequently occur in real-world scenarios: brightness, dust, defocus blur, fog, frost, and snow. The severity of corruption is varied on a scale of Level 1 (lowest) to Level 5 (highest), in steps of 1. Figure \ref{fig:resnet} and \ref{fig:vgg} show the increment in adversarial accuracy post-NSA. The mean adversarial accuracy across all severity levels (with or without NSA) is stated beside the legend and is denoted by $u$. 

As expected, we observe decreasing accuracy as the severity of noise increases. Interestingly, we notice a similar trend across both models for all corruptions. Among the corruptions, we found that the brightness does not degrade the prediction compared to other corruptions. On the other hand, the severe dust levels have profound implications. Especially for higher severity (L4 and L5), we notice that even NSAs are unable to improve the robustness. One possible reason could be that the intense dust in the scene might hamper the minimum required visibility of the NSA, leading to its ineffectiveness. However, it is essential to note that in such instances, even the target object might not be clearly visible; hence, the adversarial accuracy is lower. For corruptions like fog, frost, and snow, we see an improvement of around 12\% on average for both models. Overall, we observe that the inclusion of NSAs has a positive impact on the model's prediction. 

\subsection{Experiment III: Variation in the performance of NSA across different target classes}

In this experiment, we investigated how NSA helps to improve accuracy across different target classes. It is essential to understand the ease of artifact training for each class. The NSA is currently trained with the same hyperparameters for all classes. Table \ref{tab:tab2} gives an idea about the classes for which we may need to change the hyperparameters to achieve better performance. For such classes, we need to design powerful NSAs by either increasing the epochs during training or placing numerous and larger artifacts in the scene if possible. In the table, the natural accuracy is based on uncorrupted images. The adversarial accuracy is calculated as the mean accuracy over corrupted images (brightness, dust, defocus, fog, frost, and snow) across all severity levels (L1-L5). 

Unlike adversarial training, where optimizing against perturbations leads to lower natural accuracy, in artifact training, the natural accuracy improves along with adversarial accuracy. For all target classes, we were able to achieve 100 \% accuracy for uncorrupted images using NSA. In the scenario against corruption, we achieved around 8\% improvement in adversarial accuracy on average. We note that the NSAs could not boost the prediction for \texttt{cassette player}, indicating the need for stronger artifacts for this class. However, we want to point towards the classes like \texttt{english springer} and \texttt{chain saw} for which we improved by up to 19\%. Observing the variance in class-level performance is expected as each class has a different underlying feature. Hence, we suggest using custom hyperparameters for artifact training for each class.

\begin{table}[t]
\centering
\caption{\small
Experimental results of class-level NSA influence on ResNet18 and VGG16 model. The $\texttt{ORI}$ represents the Original image without any artifact and $\texttt{NSA}$ is the image with an artifact applied. 
}\label{tab: results_overall}
\scalebox{0.87}{
\begin{tabular}{|l|cc|cc|cc|cc|}
  \toprule
  \multirow{4}{*}{\bf \texttt{Class Name}} 
  & \multicolumn{4}{c|}{\multirow{2}{*}{\bf \texttt{Natural Accuracy}}} & \multicolumn{4}{c|}{\multirow{2}{*}{\bf \texttt{Adversarial Accuracy}}} \\ 
  &  \multicolumn{4}{c|}{}  &  \multicolumn{4}{c|}{}  \\
  \cline{2-9}
  & \multicolumn{2}{c|}{\multirow{2}{*}{\bf \texttt{ResNet18}}} & \multicolumn{2}{c|}{\multirow{2}{*}{\bf \texttt{VGG16}}} &  \multicolumn{2}{c|}{\multirow{2}{*}{\bf \texttt{ResNet18}}} & \multicolumn{2}{c|}{\multirow{2}{*}{\bf \texttt{VGG16}}}\\
  &  \multicolumn{2}{c|}{}  &  \multicolumn{2}{c|}{}  & \multicolumn{2}{c|}{}  &  \multicolumn{2}{c|}{} \\
  & \texttt{ORI} & \texttt{NSA} & \texttt{ORI} & \texttt{NSA} & \texttt{ORI} & \texttt{NSA} & \texttt{ORI} & \texttt{NSA} \\
  \midrule
  English Springer & 97 & 100 & 97 & 100  & 56 & 65 & 58 & 76\\
  French Horn & 98 & 100 & 98 & 100  & 63 & 76 & 64 & 78 \\
  Cassette Player & 99 & 100 & 99 & 100  & 77 & 78 & 81 & 81 \\
  Chain Saw & 98 & 100 & 98 & 100  & 71 & 87 & 57 & 78 \\
  Church & 98 & 100 & 99 & 100  & 75 & 80 & 81 & 85 \\
  Garbage Truck & 100 & 100 & 100 & 100  & 67 & 73 & 79 & 82\\
  Gas Pump & 96 & 100 & 96 & 100  & 48 & 57 & 53 & 64\\
  Golf Ball & 100 & 100 & 99 & 100  & 74 & 82 & 84 & 90\\
  Parachute & 98 & 100 & 98 & 100  & 93 & 97 & 87 & 96\\
  Tench & 100 & 100 & 100 & 100  & 75 & 84 & 73 & 84\\
  \rowcolor{lightgray} \bf OVERALL & 98 & \color{red}\textbf{100} & 98 & \color{red}\textbf{100} & 70 & \color{red}\textbf{77} & 72 & \color{red}\textbf{81} \\
  \bottomrule
\end{tabular}
\label{tab:tab2}
}\\\vspace{-.5em}
\end{table}

\subsection{Experiment IV: Visualizing the impact of NSAs}

\begin{figure*}[t]
     \centering 
     \includegraphics[scale=0.31]{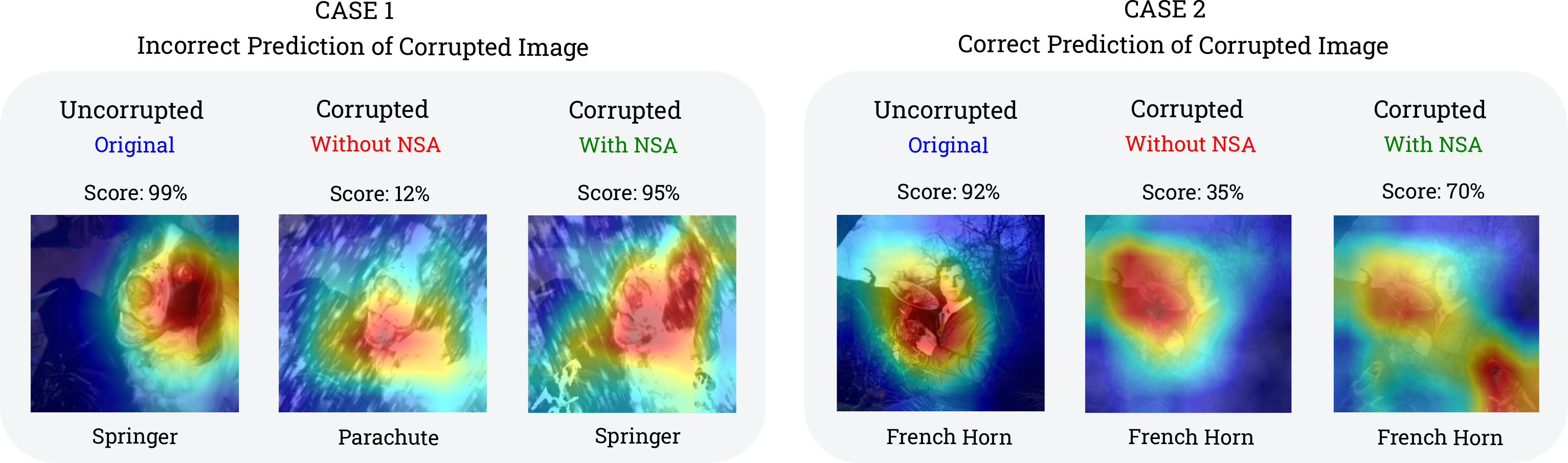}
     \caption{Saliency map visualization showing ResNet18’s perception of NSA applied unadversarial example under two scenarios.}
     \label{fig:gradcam}
 \end{figure*}

For a qualitative understanding of NSA's impact, we used a saliency map generation technique called Grad-CAM \cite{selvaraju2017grad}. We observed that there are two main scenarios (Case 1 and Case 2) where the contribution of an NSA needs to be evaluated, as shown in Figure \ref{fig:gradcam}. First, when the model performs an incorrect prediction over the corrupted image which is later rectified by placing NSAs in the scene. Second, when the original prediction is correct for the corrupted image, and the NSA assists in boosting the model's confidence in its prediction. We randomly select fog corruption for this study.

In Case 1, we see that the model focuses on the eyes, nose, mouth, and ears while making the prediction of \texttt{English Springer}. However, with the corruption, the model's focus is switched to only the mouth, leading to an incorrect prediction as \texttt{Parachute}. If we place an NSA in the scene, we observe that the model's focus remains intact on features similar to that of the original image. Interestingly, even though the NSAs are trained to increase robustness, the model does not focus on them. The NSAs help the model focus on the original rather than the artifact features. 

In Case 2 we see that the model's focus is similar in both corrupted and uncorrupted scenarios. However, we observe that the focus on \texttt{French Horn} is better in the uncorrupted (original) case, which is as expected and has a higher confidence score of 92\%. With the addition of NSAs in the scene we see that the score goes up from 35\% to 70\%. Unlike Case 1, we see that the model focuses on both the original features and the artifacts to improve the prediction confidence. It is necessary because the model was already focusing on the important features in the image yet had low confidence due to the noise. So it appears the NSAs trained to learn the class-specific information aid the model in decision-making. 

\section{Conclusion}
\label{sec:label}
Our work showcases that the presence of carefully formulated artifacts called NSAs can help the model in decision-making. Interestingly, unlike adversarial training, NSAs are capable of improving natural and adversarial accuracy at the same time. The idea of unadversarial examples inspires the training of NSAs, but we proposed a naturalistic and meaningful way of designing them. The approach ensures that the artifacts do not look out of place or suspicious to the human eye. We introduced the generators (from pre-trained GAN) in the artifact training framework to ensure that the artifacts retain the naturalistic and necessary patterns to assist the model in prediction. We also employ multiple generators to perform training of all artifacts simultaneously. We believe NSAs can be adopted to improve the prediction robustness.

\bibliographystyle{unsrt}  
\bibliography{arxiv}

\end{document}